\documentclass{article}
\usepackage{spconf,amsmath,graphicx}

\usepackage{multirow}
\usepackage{makecell}
\usepackage{color}
\usepackage{hyperref}
\usepackage{marvosym}

\title{SEMI-SUPERVISED RANKING FOR OBJECT IMAGE BLUR ASSESSMENT}
\name{Qiang Li$^\star$, Zhaoliang Yao$^\star$, Jingjing Wang, Ye Tian, Pengju Yang, Di Xie, Shiliang Pu \thanks{$^\star$Authors contribute equally to this research.} }
\address{Hikvision Research Institute, Hangzhou, China \\
\{liqiang23\Letter,yaozhaoliang,wangjingjing9,tianye10,yangpengju,xiedi,pushiliang.hri\Letter\}@hikvision.com
}

\begin{document}
%
\maketitle
\begin{abstract}
Assessing the blurriness of an object image is fundamentally important to improve the performance for object recognition and retrieval. The main challenge lies in the lack of abundant images with reliable labels and effective learning strategies. Current datasets are labeled with limited and confused quality levels. To overcome this limitation, we propose to label the rank relationships between pairwise images rather their quality levels, since it is much easier for humans to label, and establish a large-scale realistic face image blur assessment dataset with reliable labels. Based on this dataset, we propose a method to obtain the blur scores only with the pairwise rank labels as supervision. Moreover, to further improve the performance, we propose a self-supervised method based on quadruplet ranking consistency to leverage the unlabeled data more effectively. The supervised and self-supervised methods constitute a final semi-supervised learning framework, which can be trained end-to-end. Experimental results demonstrate the effectiveness of our method.
Source of labeled datasets: \color{magenta}{https://github.com/yzliangHIK2022/SSRanking-for-Object-BA}
\end{abstract}

\begin{figure}[th]
	\centering
	\includegraphics[width=0.75\linewidth]{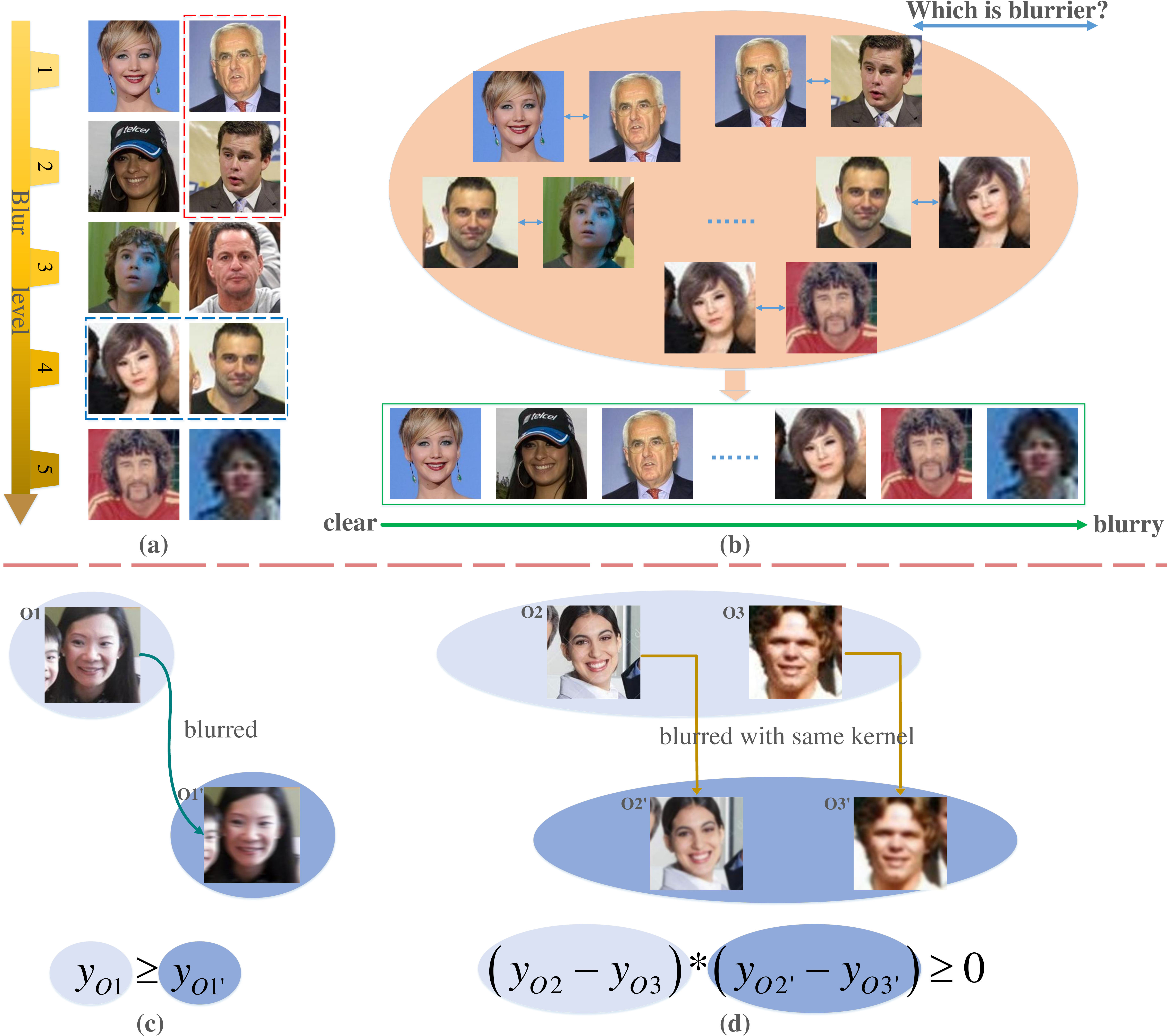}
	\caption{Different dataset constructions (a \emph{vs} b) and self-supervised learning (SSL) techniques (c \emph{vs} d): (a) label with limited quality levels; (b) label with pairwise blur orders; (c) pairwise ranking consistency for SSL; (d) our proposed quadruplet ranking consistency for SSL.}
	\vspace{-7pt}
	\label{fig:abstract}
\end{figure}

\begin{figure*}[t]
	\centering
	\includegraphics[width=0.67\linewidth]{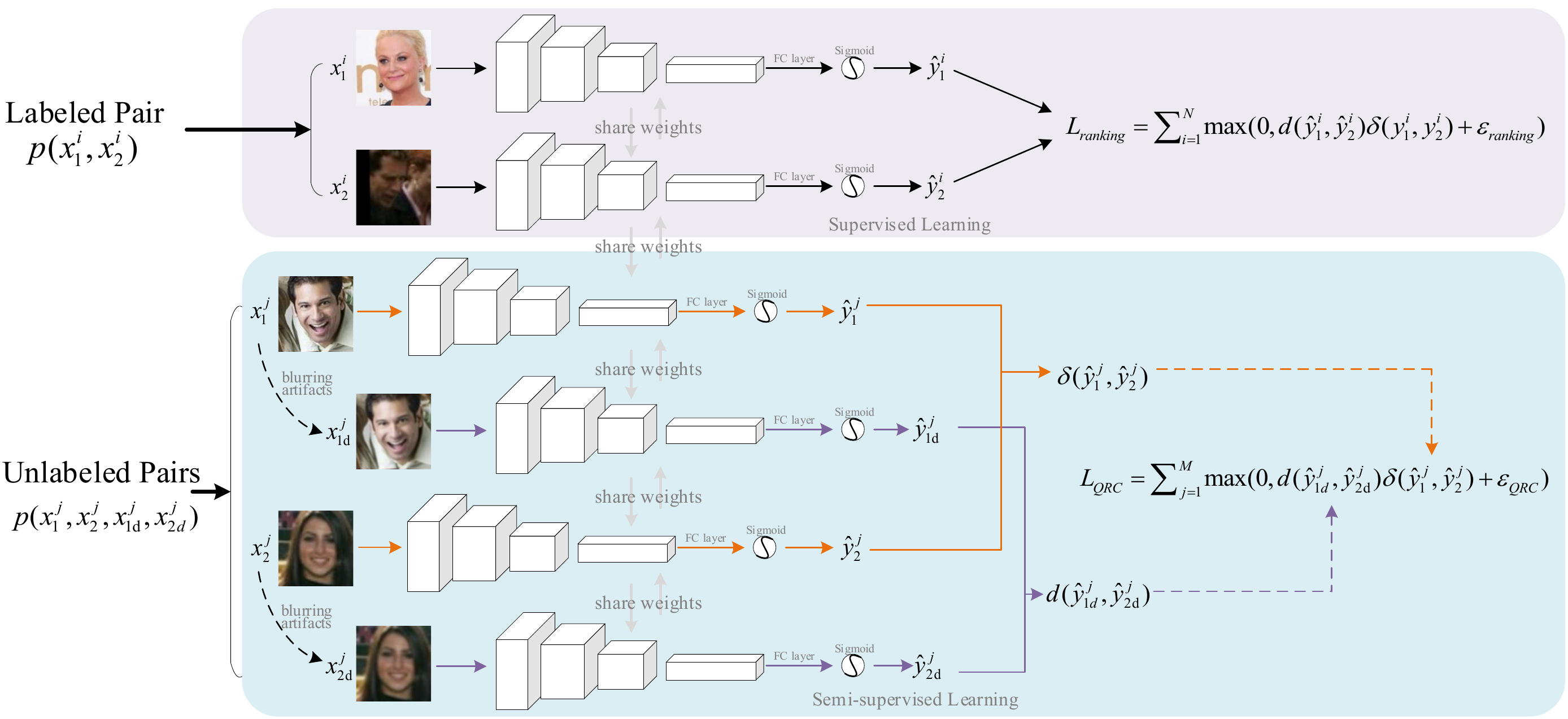}
	\vspace{-5pt}
	\caption{Overview of our approach which is consisted of supervised learning branch and self-supervised learning branch.}
	\vspace{-5pt}
	\label{fig:framework}
\end{figure*}

\section{Introduction}
\label{sec:intro}
In real-life object recognition systems (e.g. face recognition \cite{2014DeepFace}, vehicle re-identification \cite{2020Vehicle}, license plate recognition \cite{2018LPRNet}), object images usually contain one centering object and appear with different qualities. Practically, it's very important to indicate the object image quality for it can significantly improve the application experience if we can filter out low-quality object image frames. Recently many works \cite{2020Object, 2021MagFace,2021EQFace} turn to utilize the quality (e.g. occlusion, blur, illumination) of object images to improve the application performance.

As a key factor of the image quality \cite{2010A,2017Image,2018Multiscale}, object image blur assessment (Object-BA) is not well studied yet and thus our work focuses on this issue. Object-BA can be defined as: for an image that only contains one object to be recognized, the blur level of foreground object requires to be assessed with reasonable blur scores by determining how well the object image is suitable for recognition . Object-BA only evaluates the foreground object, and it is different from IQA problems since IQA attempts to assess the quality/blur of the entire image in which none or multiple objects may exist. Object-BA is also different with the object quality assessment \cite{2020Object}, since it focuses on the global quality assessment which may lack of interpretation, while we aim to decompose the factors affecting object image quality, and solve one of the most important factors assessment, i.e. the object image blur assessment.

Although convolutional neural networks have achieved great success on many visual tasks, their application on quality assessment still suffers from two challenges: lack of abundant images with reliable labels and more effective learning strategy for fine-grained quality assessment.
Current image quality assessment datasets either label the image with limited quality levels or synthesize images with different qualities using certain image degradation methods. For examples, BCNet \cite{2017Image} establishes a real-world dataset including 2000 images labeled with only blur or not, which is not sufficient for fine-grained quality assessment. To solve this problem, \cite{2017NIMA} labels the images with ten quality levels. However, labeling images with accurate quality levels is very difficult for humans, which may limit the number of quality levels, and lead to confusing labels between inter and intra levels. Overlaps of different blur levels are shown as red dotted boxes in Fig.\ref{fig:abstract} (a), and even within the same level, the blur of two images can be distinct (blue dotted box in Fig.\ref{fig:abstract} (a)). Others resort to synthesized images to alleviate this problem. LIVE \cite{2003LIVE} and TID2008 \cite{2009TID2008} synthesize large amount of distorted images from un-distorted images and the blur level is determined by different blur kernels. However there is a large domain gap between synthesized images and real ones, which may lead to severe performance degradation in real-world applications.

For more effective learning strategy, limited by current datasets, most existing methods \cite{2017Image, 2018Multiscale, 2020Defocus} treat quality assessment as a classification task or directly regress the true quality levels. These methods have limited performance in real applications due to the aforementioned reasons. Some researchers \cite{2017RankIQA, 2017dipIQ} convert it as a learn-to-rank problem according to their quality levels or based on synthesized images. However they are only based on pairwise relationships between the original and corresponding synthesized degraded images, which limits the representation ability of the learned features.

To solve the above limitations, we firstly proposed to label the object image qualities with pairwise ranks, as labeling which image is sharper than the other is much more easier than labeling blur levels. Therefore, we construct a new object-BA dataset (named FIB) by hiring crowdsourcing annotators to judge the blur ranking of all image pairs from two popular face datasets \cite{2008Labeled,2014Learning}. In this way, we can get the total ranking of all images as shown in Fig.\ref{fig:abstract} (b). The ranking labels are all from real images rather than the synthesized images \cite{2003LIVE, 2009TID2008}, which would benefit the applications in real world. Based on this well-labeled dataset, to get fine-grained quality scores, we propose to learn the scores only regularized by the ranking relationships. To further improve the performance, we propose a semi-supervised learning framework to leverage the large-scale unlabeled data \cite{2016WiderFace,2015MegaFace} (named as FIB-unlabeled). The framework is shown in Fig.\ref{fig:framework}, which consists of a supervised learning branch as mentioned above and a self-supervised learning branch. In the self-supervised learning branch, we propose to construct a quadruplet instead of an ordinary pair as previous methods \cite{2017RankIQA, 2017dipIQ} to generate rank relationships in a self-supervised manner. In a quadruplet, two images are blurred to generate two corresponding synthesized images using a same blur kernel. Therefore, the rank relationships of the two synthesized images is consistent with the original ones as shown in Fig.\ref{fig:abstract} (d). While in an ordinary pair, only one image is blurred and the rank relationship which can be leveraged is that the rank of the blurred image is lower than the original image as shown in Fig.\ref{fig:abstract} (c). The two images in a quadruplet  are different from each other, containing different contents, backgrounds, illumination, occlusion etc, while the only difference in an original pair is the blur degree. Therefore, learning with quadruplets can make the learned feature more robust against various changes, which can lead to better performance for real-world applications.

Our contributions are as follows:
(1) We define a new object image blur assessment problem, which are valuable for interpretable object image quality assessment, and establish a large-scale realistic dataset using pairwise ranks for reliable labeling to assess the object image blurriness which would benefit the research of object image quality assessment in real-life applications.
(2) We propose a new method to get absolute fine-grained blur scores only based on pairwise ranks, and a new semi-supervised framework based on quadruplet inputs which can leverage large amounts of unlabeled images more efficiently for better performance.

\begin{figure}[t]
	\centering
	\includegraphics[width=0.7\linewidth]{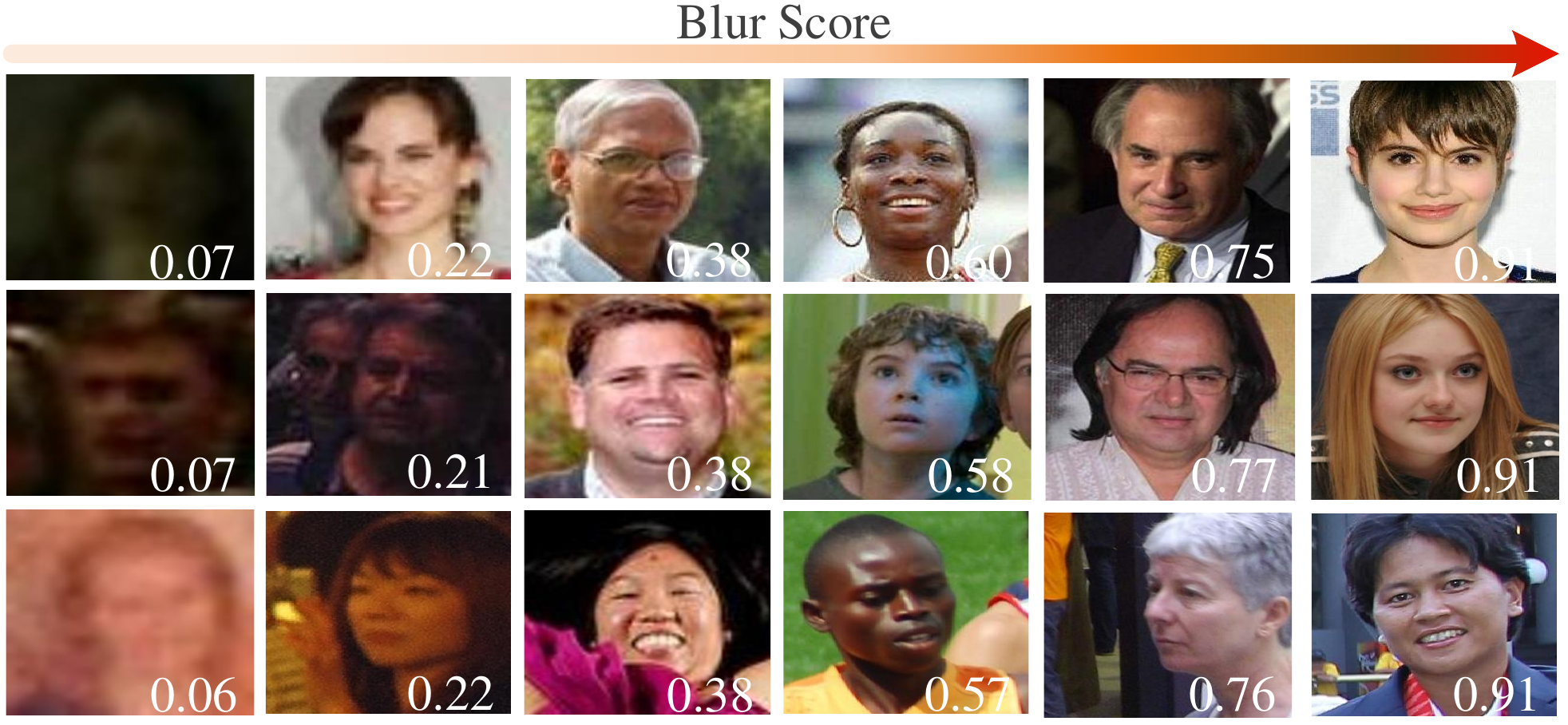}
	\caption{
		Examples of prediction scores on Test1(first row), Test2(second row) and Test3(third row).
		Blur scores are written in white color under image and sorted from low to high.
	}
	\label{fig:ranking_examples}
\end{figure}

\begin{figure}[t]
	\centering
	\includegraphics[width=0.8\linewidth]{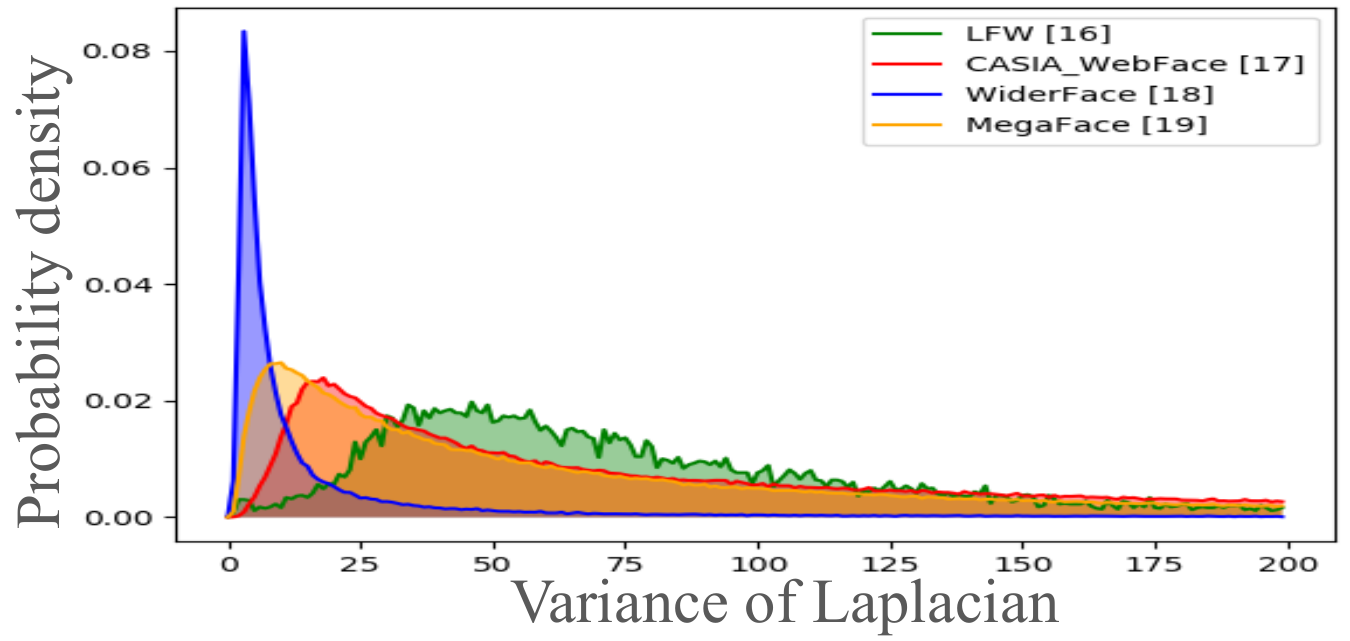}
	\vspace{-10pt}
	\caption{
		Blur distributions of different face datasets. We use Laplacian operator to filter face images and take the variance of filtered images as their blur level roughly..
	}
	\vspace{-5pt}
	\label{fig:blur_statistics}
\end{figure}

\section{DataSet Construction}
\label{sec:face image blur data}
Different from previous datasets which label images with limited quality levels or use synthesized images, we aim to construct a large-scale realistic dataset with reliable label for fine-grained object image blurriness assessment.
In this paper, we focus on face image blurriness assessment. To enrich the dataset, we sample images of different blur levels randomly from LFW \cite{2008Labeled} and CASIA-WebFace \cite{2014Learning} to construct our supervised training dataset FIB. The detailed labeling steps are as follows. Firstly, faces are cropped and resized to 256$\times$256 resolution. Secondly, image pairs are randomly sampled for labeling. Due to the huge labeling cost for labeling all the image pairs, we propose to label randomly sampled image pairs to cover more images. Thirdly, image pairs are labeled by three annotators and the label is determined by the majority voting strategy. Specifically, letting $x^{i}_{1}, x^{i}_{2}$ denote the two images in an image pair and $y^{i}_{1}, y^{i}_{2}$ denote their corresponding blur scores, the relative ranking label of an image pair $p(x^{i}_{1}, x^{i}_{2})$ is labeled as $\delta(y^{i}_{1}, y^{i}_{2}){=}{-}1$ when the majority of annotators agree $x^{i}_{1}$ is blurrier than $x^{i}_{2}$, and $\delta(y^{i}_{1}, y^{i}_{2}){=}{+}1$ otherwise.
In this way, we get a dataset with 10,000 labeled image pair, which covers 13,751 images and is named FIB-10K. To further investigate the effect of the label number, we randomly sample half of the labeled image pairs, and construct a sub-dataset named FIB-5K.

Although we construct the largest realistic dataset compared with other quality assessment datasets, the image number of the dataset is still limited. To alleviate this, we introduce a large-scale unlabeled data. To simulate the real scenarios where unlabeled data may come from a datset with different distribution from the labeled one, we randomly crop faces from WiderFace \cite{2016WiderFace} and construct a unlabeled dataset FIB-unlabeled which covers 86,065 images.

To fully evaluate the effectiveness of the quality assessment algorithms, we construct three test datasets for evaluating them under different settings. The first one is the intra-dataset testing, in which test data is sampled exclusively from LFW \cite{2008Labeled} and CASIA-WebFace \cite{2014Learning} as the labeled data. The second one is the half-inter-dataset testing, in which test data is sampled exclusively from WiderFace \cite{2016WiderFace} as the unlabeled data. The third one is the inter-dataset testing, in which test data is sampled from a new dataset MegaFace \cite{2015MegaFace} (the blur distributions of these datasets are shown in Fig.\ref{fig:blur_statistics}). We denote the above datasets as Test1, Test2 and Test3 respectively. The labeling steps are similar as the supervised training dataset FIB. The only difference is that, we label all the image pairs to get the total ranking order in each dataset. Due to the huge labeling cost, each of the test datasets only has 100 images, however the number of the labeled pairs is 4,950. Some examples of testing datasets are shown in Fig.\ref{fig:ranking_examples}. The statistics of our datasets are shown in Tab.1.

\begin{table}[t]
	\caption{Statistics of training datasets FIB-labeled and FIB-unlabeled and testing datasets Test1, Test2 and Test3.}
	\begin{center}
      \small
		\begin{tabular}{|p{2.0cm}<{\centering}| p{1.0cm}<{\centering}| p{2.25cm}<{\centering}| p{1.5cm}<{\centering}|}
			\hline
			Dataset  & Source & Labeled pairs No. & Images No. \\
			\hline\hline
			FIB-5k & \cite{2008Labeled,2014Learning} & 5,000 & 8,093 \\ \hline
			FIB-10k & \cite{2008Labeled,2014Learning} & 10,000 & 13,751 \\ \hline
			FIB-unlabeled & \cite{2016WiderFace} & - & 86,065  \\ \hline
			Test1 & \cite{2008Labeled,2014Learning} & 4,950 & 100 \\ \hline
			Test2 & \cite{2016WiderFace} & 4,950 & 100 \\ \hline
			Test3 & \cite{2015MegaFace} & 4,950 & 100 \\ \hline
		\end{tabular}
	\end{center}
	\label{table:dataset}
	\vspace{-15pt}
\end{table}

\section{Proposed Approach}
\label{sec:method}
The framework is shown in Fig.\ref{fig:framework}. It consists of a supervised learning branch which leverages the pairwise labeled data to predict blur scores, and a self-supervised branch which leverages the unlabeled data to improve the generalization performance. The details are described in following subsections.

\subsection{Supervised Learning Based on Pairwise Ranking}
\label{sec:super_ranking}
The supervised learning branch takes an image pair $p(x^{i}_{1}, x^{i}_{2})$ with relative ranking label $\delta(y^{i}_{1}, y^{i}_{2})$ as input and predicts blur scores $\hat{y}^{i}_{1}$, $\hat{y}^{i}_{2}$.
Each sub-network of this branch consists of a backbone network, followed by dense layers and finally a sigmoid to get the blur scores. As we don't have the true blur scores, we propose to use a pairwise margin ranking loss to regularize the learned blur scores, which is defined as:

\begin{equation}
\small
L_{ranking}{=}{\frac{1}{N}}\sum^{N}_{i=1}max(0, d(\hat{y}^{i}_{1},\hat{y}^{i}_{2})*\delta(y^{i}_{1}, y^{i}_{2}) {+} \varepsilon_{ranking})
\label{eq:hinge_loss}
\end{equation}
where $\varepsilon_{ranking}$ is a controllable ranking margin; $N$ is the number of image pairs $p(x^{i}_{1}, x^{i}_{2})$;  $d(\hat{y}^{i}_{1}, \hat{y}^{i}_{2})$ denotes the distance of blur scores which is $d(\hat{y}^{i}_{1}, \hat{y}^{i}_{2}){=}({-}1){*}(\hat{y}^{i}_{1}{-}\hat{y}^{i}_{2})$.

During testing, we use one of the sub-network to predict the blur score of the input image. Some predicted results are shown in Fig.\ref{fig:ranking_examples}, from which we can see our method can predict reasonable blur scores with only relative ranking labels.

\subsection{Self-supervised learning Based on Quadruplet Ranking Consistency}
\label{sec:semi_super_ranking}
Based on an observation that the relative ranking relationship of two images would not change after they have been subjected to the same image distortion attack, we propose to use a quadruplet instead of an ordinary pair to generate rank relationships in a self-supervised manner due to the reasons analyzed in the introduction section.
Specifically, let $x^{j}_{1d}, x^{j}_{2d}$ denote the corresponding blurred images of $x^{j}_{1}, x^{j}_{2}$ with the same blur kernel. Then the relative ranking label $\delta(\hat{y}^{j}_{1d}, \hat{y}^{j}_{2d})$ of $x^{j}_{1d}, x^{j}_{2d}$ should be the same as the one $\delta(\hat{y}^{j}_{1}, \hat{y}^{j}_{2})$ of $x^{j}_{1}, x^{j}_{2}$, and $x^{j}_{1}, x^{j}_{2}, x^{j}_{1d}, x^{j}_{2d}$ form a quadruplet.
The self-supervised learning branch takes an image quadruplet as input, and we propose a Quadruplet Ranking Consistency (QRC) loss for training which is defined as:
\begin{equation}
\small
L_{QRC}{=}{\frac{1}{M}}\sum^{M}_{j=1}max(0, d(\hat{y}^{j}_{1d}, \hat{y}^{j}_{2d})*\delta(\hat{y}^{j}_{1}, \hat{y}^{j}_{2}) {+} \varepsilon_{QRC}).
\label{eq:bccr_form}
\end{equation}
where $\varepsilon_{QRC}$ is a controllable ranking margin; $M$ is the number of image quadruplets.

\section{EXPERIMENTS}
\label{sec:exps}

\subsection{Implementation Details}
\label{sec:implementation}
We choose ResNet-18 \cite{2016resnet} as our network backbone for balancing performance and efficiency.
The input images are center-cropped from original images, and resized to  224$\times$224.
At training stage, batch size is set to 60, and SGD is used to optimize the model with weight decay $0.0005$ and moment $0.9$.
Learning rate is initialized from 0.001 and scheduled by a CosineAnnealingLR policy \cite{2019pytorch}.
We use
\textit{Spearman Rank Order Correlation Coefficient (SROCC)}
as our evaluation protocol like others \cite{2017RankIQA, 2019Regression,2017dipIQ}.

\begin{table}[t]
	\renewcommand\arraystretch{1.0}
	\setlength\tabcolsep{4.2pt}
	\caption{Experimental results.}
	\begin{center}
		\begin{tabular}{|p{1.3cm}<{\centering}| p{2.85cm}| p{0.9cm}<{\centering}| p{0.9cm}<{\centering}|p{0.9cm}<{\centering}|  }
			\hline
			Trainset  & Method & Test1 & Test2 & Test3 \\
			\hline\hline
			\multirow{4}{*}{ FIB-10K } & Baseline & 0.9408 & 0.9643 & 0.9258 \\ \cline{2-5}
			& Ours\_RankIQA \cite{2017RankIQA} & 0.9286 & 0.9704 & 0.9275  \\ \cline{2-5}
			& Ours\_LSEP \cite{2019Regression} & 0.9274 & 0.9605 & 0.9210 \\ \cline{2-5}
			& Ours & \bf 0.9516 & \bf 0.9734 & \bf 0.9430 \\ \cline{1-5}
			\multirow{4}{*}{ FIB-5K } & Baseline & 0.9285 & 0.9572 & 0.9091\\ \cline{2-5}
			& Ours\_RankIQA \cite{2017RankIQA} & 0.9240 & 0.9633 & 0.9125 \\ \cline{2-5}
			& Ours\_LSEP \cite{2019Regression} & 0.9180 & 0.9534 & 0.9005 \\ \cline{2-5}
			& Ours & \bf 0.9480 & \bf 0.9723 & \bf 0.9330 \\ \cline{1-5}
		\end{tabular}
	\end{center}
	\label{table:ranking_performace}
	\vspace{-15pt}
\end{table}

\subsection{Experimental Results}
\label{sec:ranking_performance}
Since previous methods are designed to regress the quality scores limited by existing datasets, while our dataset is labeled with pairwise rank relationships without the true quality scores. Therefore, directly comparing our method with them on our dataset is impossible. We compare our methods with different variations to show the effectiveness of the proposed semi-supervised framework and the quadruplet based ranking consistency loss. We denote baseline as our method with only the supervised learning branch. We denote ours-RankIQA and ours-LSEP as our method with the quadruplet based ranking consistency loss in the self-supervised branch replaced by the rank loss proposed in RankIQA \cite{2017RankIQA} and LSEP \cite{2019Regression}. It is worth noting that both of the two rank losses use the pairwise rank relationships, i.e. the rank of the blurred image should be lower than the original image.

The experimental results are shown in Tab.\ref{table:ranking_performace}. It can be seen that Ours\_RankIQA and Ours\_LSEP achieve slightly worse performance on Test1 and comparable performance on Test2 and Test3. It shows that only leveraging the pairwise rank relationships in the self-supervised branch would hinder the performance under intra-dataset testing and would not improve the performance under half-inter-dataset testing and inter-dataset testing, since the image pairs only include blur changes which is easy to be learned by the network. Since the proposed quadruplet can introduce various changes during the feature learning, our method outperforms the others significantly under all of the three test settings, which verifies the effectiveness the proposed semi-supervised framework and the quadruplet based ranking consistency loss.
Moreover, comparing the performances with FIB-5K and FIB-10K as the supervised dataset, the performance of other methods drop obviously with less labeled data, while the performance of our method is comparable under intra-dataset and half-inter-dataset testings, and only slightly worse under inter-dataset testing with only half of the labeled data. It shows that our method can alleviate the lack of labeled data by effectively leveraging the unlabeled data.

Furthermore, we illustrate the blur scores learned by our method in Fig.\ref{fig:ranking_examples}, from which we can see our method can predict reasonable and discriminate score for blur assessment, although learned with only pairwise rank labels.

\section{CONCLUSION}
\label{sec:conclusion}

In this paper, we define a new object image blur assessment problem, and establish a large-scale realistic dataset with pairwise rank labels.
Based on this dataset, we proposed a new semi-supervised learning method which consists of a supervised learning branch and a self-supervised learning branch.
The supervised learning branch can learn to predict the fine-grained blur scores with only pairwise rank labels. The self-supervised learning branch can leverage the unlabeled data to improve the performance. To achieve this goal, a quadruplet ranking consistency loss is proposed to make the learned feature more robust against various changes.
Extensive experiments under intra-dataset, half-inter-dataset and inter-dataset settings show the effectiveness the proposed semi-supervised framework and the quadruplet based
ranking consistency loss.
\vfill\pagebreak
\bibliographystyle{IEEEbib}
\bibliography{refs}

\end{document}